\documentclass[10pt,twocolumn,letterpaper]{article}

\usepackage[final]{cvpr}      

\usepackage{graphicx}
\usepackage{amsmath}
\usepackage{amssymb}
\usepackage{booktabs}

\usepackage[pagebackref,breaklinks,colorlinks]{hyperref}

\usepackage[capitalize]{cleveref}
\crefname{section}{Sec.}{Secs.}
\Crefname{section}{Section}{Sections}
\Crefname{table}{Table}{Tables}
\crefname{table}{Tab.}{Tabs.}

\def\cvprPaperID{*****} 
\def\confName{CVPR}
\def\confYear{2022}

\begin{document}


\title{Multiple Object Tracking Challenge Technical Report for Team MT$\_$IoT}

\author{Feng Yan, \enspace Zhiheng Li, \enspace Weixin Luo, \enspace Zequn jie, \enspace Fan Liang, \enspace Xiaolin Wei, \enspace Lin Ma\\
\normalsize Meituan Inc.\\
{\tt\small {\{yanfeng05,
lizhiheng03,
luoweixin,
jiezequn, liangfan02, weixiaolin02,
malin11\}}@meituan.com}}

\maketitle

\begin{abstract}

This is a brief technical report of our proposed method for Multiple-Object Tracking (MOT) Challenge in Complex Environments. In this paper, we treat the MOT task as a two-stage task including human detection and trajectory matching. Specifically, we designed an improved human detector and associated most of detection to guarantee the integrity of the motion trajectory. We also propose a location-wise matching matrix to obtain more accurate trace matching. Without any model merging, our method achieves $66.67^{2}$ HOTA and $93.97^{1}$ MOTA on the DanceTrack challenge dataset. 

\end{abstract}

\section{Method}
\subsection{Overview of the proposed method}

\begin{figure*}
\centering
\includegraphics[width=1.0\textwidth]{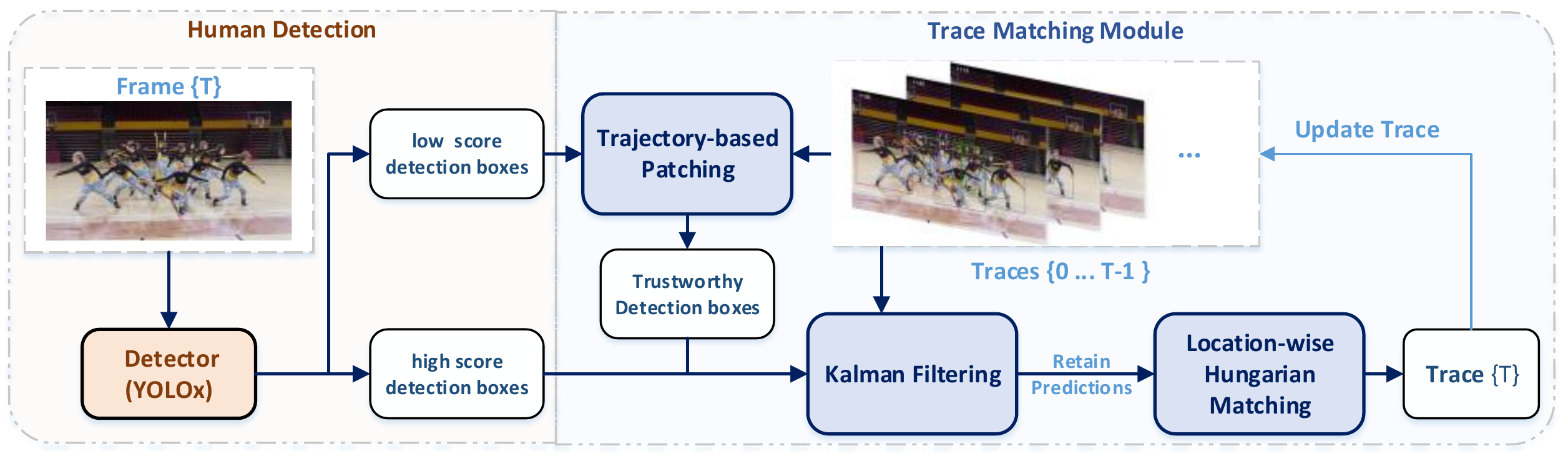}
\caption{Overall architecture of our proposed method.}
\label{fig:architecture}
\end{figure*}

The schematic diagram of out method is illustrated in Fig.~\ref{fig:architecture}. We presented a two-stage tracking-by-detection structure including human detector and trace matching module. For the detection phrase, we adopted the YOLOX~\cite{ge2021yolox} for the object-intensive detection task. Then the boxes with low scores are fed into the Trajectory-based Patching module to determine which 
detection can be retained. All trustworthy detection will be matched with existing traces subsequently.

\subsection{Human Detector}
We adopted the YOLOX~\cite{ge2021yolox} as the human detector. For the tracking task, most false detection are caused by occlusion between targets in multi-object scenarios. To solve this problem, we introduce two public datasets Crowdhuman~\cite{shao2018crowdhuman} and MOT20~\cite{dendorfer2020mot20} to enhance the robustness of the detector for multi-object scenes. After pretrained on both datasets, the human detector greatly improved the detection of obscured targets, further enhancing the continuity of the motion trajectory.

\subsection{Trace Matching Module}

\paragraph{Trajectory-based Patching Method} Inspired by Byte Track~\cite{zhang2021bytetrack} and OC-Sort~\cite{cao2022observation}, we believe detection boxes with low scores also contribute to securing the continuity of trajectory. We proposed a trajectory-based patching method to capture useful detection boxes from ones with low confidence scores. The method first utilize boxes with high scores to match with existing trajectories. Then for the unmatched traces, we try to correlate the last detection of each trajectory with the remaining boxes with low scores. Specifically, the method calculates IoU(Intersection over Union) scores between the last detection of each trace and each box with low score, then the box with highest IoU score is considered to be the best match. If there still exist unmatched traces, the prediction from the Kalman filtering will be retained as a short-term observation. Furthermore, we also impose the limitation to existing traces. When the confidence score is lower than the threshold, the trajectory will be considered to be untrustworthy and won't be correlated with any box in current observation. In most scenarios, occlusion is often accompanied by a slow decrease in detection score. The proposed patching method helps to maintain the trace coherence in challenging situations such as target occlusion.

\paragraph{Location-wise Matching Matrix} In the dancing scene, the height of dancers change noticeably with the distance away from the camera. The height change in successive frames can provide clues to the dancer's location, and helps to remove unreasonable matches caused by overlapping targets. For the Hungarian matching algorithm, each element of matching matrix represents the distance calculated based on the IoU score between each trajectory and detection box. In the dance tracking task, we utilize the height of detection box to calculate the cost distance. Since the height of the human varies slightly between adjacent observations, the location-wise matching matrix 
drives the trajectory to find the best match among boxes of similar height. In addition, the horizontal movement of the dancer is usually violent than the vertical movement. Calculating distances based on height can smooth out the noise caused by rapid movement.

\section{Experiment Results}

In this section, we present the results of ablation experiments for the pretrained human detector, trajectory-based patching method and location-wise matching matrix. We also display two unproductive cases with corresponding explanations, which may contribute to tracking tasks in other scenarios.

\subsection{Human Detector Pre-Training}

We introduced two public datasets Crowdhuman~\cite{shao2018crowdhuman} and MOT20~\cite{dendorfer2020mot20} to boost the performance of the human detector in occlusion scenes. After pre-training, the baseline is trained and tested on DanceTrack dataset as shown in {TABLE~\ref{table:pretrain}}. It can be seen from the table that pre-training provides a significant improvement in the tracking performance of the model.

\setlength{\tabcolsep}{4pt}
\begin{table}
\begin{center}
\caption{Ablation study for pretraining on different datasets, where Crowd and MOT presents Crowdhuman and MOT20 dataset respectively. }
\label{table:pretrain}
\begin{tabular}{lccccc}
\hline\noalign{\smallskip}
Datasets & HOTA & DetA & AssA & MOTA & IDF1\\
\noalign{\smallskip}
\hline
\noalign{\smallskip}
- & 55.41 & 80.99 & 38.04 & 91.80 & 55.47\\
Crowd & 57.76 & 83.27 & 40.19 & 93.36 & 57.15\\
Crowd + MOT & 58.82 & 83.33 & 41.63 & 93.44 & 58.93\\
\hline
\end{tabular}
\end{center}
\end{table}
\setlength{\tabcolsep}{1.4pt}

\subsection{Trajectory-based Patching Method}

In this section, we present the experiment results of the proposed Trajectory-based Patching method. We also conducted the ablation study for different IoU algorithms including D-IoU~\cite{zheng2020distance}, G-IoU~\cite{rezatofighi2019generalized} and C-IoU~\cite{zheng2020distance} as shown in TABLE~\ref{table:iou}. 

\setlength{\tabcolsep}{4pt}
\begin{table}
\begin{center}
\caption{Ablation study for the proposed Trajectory-based Patching(TP) method with different IoU algorithms.}
\label{table:iou}
\begin{tabular}{lccccc}
\hline\noalign{\smallskip}
Datasets & HOTA & DetA & AssA & MOTA & IDF1\\
\noalign{\smallskip}
\hline
\noalign{\smallskip}
w/o TP & 58.82 & 83.33 & 41.63 & 93.44 & 58.93\\
w/ TP(DIoU) & 60.13 & - & - & - & -\\
w/ TP(GIoU) & 60.12 & - & - & - & -\\
w/ TP(CIoU) & 60.36 & 82.96 & 44.00 & 93.95 & 64.00\\
\hline
\end{tabular}
\end{center}
\end{table}
\setlength{\tabcolsep}{1.4pt}

\subsection{Location-wise Matching Matrix}

We compare the effects of different matching matrices on tracking performance as shown in TABLE~\ref{table:local-wise}. We conducted the ablation experiments with area-based and height-based IoU scores, respectively. As shown in the table, the proposed location-wise matching method provides a significant improvement in both tracking and detection performance on the DanceTrack dataset.

\setlength{\tabcolsep}{4pt}
\begin{table}
\begin{center}
\caption{Ablation study for different matching matrix. }
\label{table:local-wise}
\begin{tabular}{lccccc}
\hline\noalign{\smallskip}
Cost Matrix & HOTA & DetA & AssA & MOTA & IDF1 \\
\noalign{\smallskip}
\hline
\noalign{\smallskip}
Area-based IoU & 60.36 & 82.96 & 44.00 & 93.95 & 64.00\\
Height-based IoU & 66.66 & 84.14 & 52.95 & 93.97 & 70.60\\
\hline
\end{tabular}
\end{center}
\end{table}
\setlength{\tabcolsep}{1.4pt}

\subsection{Unproductive cases}

In this section, we present two attempts to boost the tracking performance including BoT-Sort~\cite{aharon2022bot} and model ensemble.

\paragraph{BoT-Sort} BoT-Sort proposes the camera-motion compensation(CMC) to alleviate the effects of tracker movement. The CMC obtains the camera transformation matrix by aligning adjacent frames to achieve more accurate foreground and background matching. However, we found that considering camera motion slightly degrade the performance in dance tracking task. For the DanceTrack dataset, most videos were captured with no obvious changes in perspective. After considering camera motion compensation, small detection offsets of overlapping targets may result in matching to the wrong trajectory, leading to a slight degradation of tracking performance.

\setlength{\tabcolsep}{4pt}
\begin{table}
\begin{center}
\caption{Ablation study for camera-motion compensation(CMC).}
\label{table:bot-sort}
\begin{tabular}{lccccc}
\hline\noalign{\smallskip}
Models & HOTA & DetA & AssA & MOTA & IDF1 \\
\noalign{\smallskip}
\hline
\noalign{\smallskip}
w/o CMC & 58.82 & 83.33 & 41.63 & 93.44 & 58.93\\
w/ CMC & 56.91 & 80.92 & 40.13 & 91.65 & 57.72\\
\hline
\end{tabular}
\end{center}
\end{table}
\setlength{\tabcolsep}{1.4pt}

\paragraph{Model Ensemble} We explored the effect of model ensemble on tracking performance. For each ensemble, We mix the tracking results of different models and filter out the repeated predictions by the Non-Maximum Suppression (NMS) algorithm. As shown in TABLE~\ref{table:model-merge}, tracking performance is not improved after model ensemble. For the dance tracking task, mix results from different models tends to add more negative samples which can't be filtered out by NMS, leading to a degradation of tracking performance.

\setlength{\tabcolsep}{4pt}
\begin{table}
\begin{center}
\caption{Ablation study for model ensemble.}
\label{table:model-merge}
\begin{tabular}{lccccc}
\hline\noalign{\smallskip}
Models & HOTA & DetA & AssA & MOTA & IDF1 \\
\noalign{\smallskip}
\hline
\noalign{\smallskip}
Model A & 55.41 & 80.99 & 38.04 & 91.80 & 55.47\\
Model B & 58.82 & 83.33 & 41.63 & 93.44 & 58.93\\
Merge AB  & 49.55 & 77.18 & 31.96 & 87.20 & 47.03\\
\hline
\end{tabular}
\end{center}
\end{table}
\setlength{\tabcolsep}{1.4pt}

\section{Conclusion}

In this paper, we presented a two-stage tracking-by-detection architecture with improved human detector for dance tracking task. We also proposed the Trajectory-based Patching method and Location-wise Matching Matrix to achieve better performance. If you want to run the detection trace model on ultra-low hardware, you can also refer to the other two articles~\cite{bai2019efficient}\cite{bai2020optimized}, which are currently used in real projects.

{\small
\bibliographystyle{ieee_fullname}
\bibliography{egbib}
}

\end{document}